\documentclass{new_tlp}
\usepackage{mathptmx}
\usepackage{times}  %Required
\usepackage{helvet}  %Required
\usepackage{courier}  %Required
\usepackage{url}  %Required
\usepackage{graphicx}  %Required
\usepackage{mathptmx}
\usepackage{amsmath}
\usepackage{mdframed}
\usepackage{amssymb}
\usepackage{multirow}
\usepackage{enumitem}
\usepackage{txfonts}
\usepackage[linesnumbered,ruled,vlined]{algorithm2e}
\usepackage{bm}
\usepackage{xcolor}
\usepackage{framed}
\usepackage{tabularx}
\usepackage{arydshln}
\usepackage{subcaption}
\newcommand\bcmdtab{\noindent\bgroup\tabcolsep=0pt}%

  \title[Theory and Practice of Logic Programming]
        {Incremental and Iterative Learning of Answer Set Programs from Mutually Distinct Examples}

  \author[Arindam Mitra and Chitta Baral]
         {Arindam Mitra and Chitta Baral\\
         Arizona State University\\
         \email{\{amitra7,chitta\}@asu.edu}}

\jdate{February 2018}
\pubyear{2018}
\pagerange{\pageref{firstpage}--\pageref{lastpage}}
\doi{S1471068401001193}

\newtheorem{definition}{Definition}

\newtheorem{theorem}{Theorem}
\begin{document}
%\label{key}\nocite{*}% includes all entries of BibTeX database into the list of references.

\label{firstpage}

\maketitle

\begin{abstract}
	
	Over the years the Artificial Intelligence (AI) community has produced several datasets which have given the machine learning algorithms the opportunity to learn various skills across various domains. However, a subclass of these machine learning algorithms that aimed at learning logic programs, namely the Inductive Logic Programming algorithms, have often failed at the task due to the vastness of these datasets. This has impacted the usability of knowledge representation and reasoning techniques in the development of AI systems. In this research, we try to address this scalability issue for the algorithms that learn answer set programs. We present a sound and complete algorithm which takes the input in a slightly different manner and performs an efficient and more user controlled search for a solution. We show via experiments that our algorithm can learn from two popular datasets from machine learning community, namely bAbl (a question answering dataset) and MNIST (a dataset for handwritten digit recognition), which to the best of our knowledge was not previously possible. The system is publicly available at \url{https://goo.gl/KdWAcV}. This paper is under consideration for acceptance in TPLP.
\end{abstract}


%\tableofcontents

\section{Introduction}
\begin{figure}
	
	\includegraphics[scale=0.6]{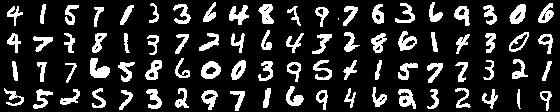}
	\caption{A set of images from the MNIST dataset.}
	\label{mnist}
\end{figure}

Answer Set Programming has emerged as a powerful tool for knowledge representation and reasoning. To use this tool for an application, however, one needs application specific knowledge. For E.g., if a system uses answer set programming to answer the question from column $1$ in Table \ref{table:dataset} the system needs to know that ``X is to the right of Y IF Y is to the left of Z and Z is above X''. Inductive Logic Programming algorithms aim to learn these kinds of knowledge from a dataset. However, existing ILP algorithms have limited scalabilty and often fail to learn knowledge from a machine learning dataset. This leads to manual construction of a knowledge base which can be very time consuming and may not be practical sometimes. For E.g., for applications where an effective representation of the rules is unknown, such as for the case of handwritten digit recognition (Fig. \ref{mnist}), one may need to try several representations before settling down for a winner. However, this may be unrealistic given that MNIST dataset (Fig. \ref{mnist}) contains $50,000$ examples and writing down the rules that explain all these examples for a particular choice of representation will take significant amount of time.

In this work, we consider this scalability issue. We observe that one major obstruction in scalability arises from the discrepancy between the definition of Inductive Logic Programming and the structure of a machine learning dataset. The learning problem in Inductive Logic Programming (ILP)  is defined as follows \cite{muggleton1991inductive}:
\begin{definition}[\bfseries{Inductive Logic Programming}]
	Given a set of positive examples $E^+$, negative examples $E^-$ and some background knowledge $B$, an ILP algorithm finds an Hypothesis $H$ such that, 
	\begin{center}
		$B \cup H \models E^+$, 
		$B \cup H  \not \models E^-$
	\end{center}
	The hypothesis space is restricted with a language bias that is specified by a series of mode declarations $M$.
\end{definition}

  A machine learning dataset on the other hand contains a series of $\langle x,y \rangle$  pairs, $x$ being the input and $y$ being the desired output (Table \ref{table:dataset}). To work with an ILP algorithm, one needs to first convert the $\langle x,y \rangle$ pairs in the format of $\langle B,E^+,E^-\rangle$. The conversion process is carried out by the user and so there might be some variations. However, normally the sets $E^+$ and $E^-$  are created using $y$'s and the $x$'s go inside $B$. Extra care is taken so that different $\langle x,y \rangle$ pairs do not interfere with each other. Table $2(a)$ shows one example of this process. 
Since the number of $\langle x,y \rangle$ pairs are usually large, the problem instance becomes too big for the ILP solvers to handle . For example, consider someone wants to employ an ILP algorithm to learn from a question answering task from bAbI dataset \cite{weston2015towards}, which contains $1,000$ comprehension examples similar to the ones in Table \ref{table:dataset}. The resulting background knowledge $B$ will contain about $10,000$ facts and $E^+$ will contain $1,000$ positive annotations pertaining to answers and $E^-$ will contain a total of $1,000$ negative examples describing what is not an answer for each question. An ILP solver such as XHAIL \cite{ray2009nonmonotonic} will throw memory errors when given an input of this size.  The question that we ask here is ``can we find  a solution to the ILP problem without considering all the $\langle x,y \rangle$ pairs together ?'' We show that the answer is yes. In fact it is possible to find a solution considering only one $\langle x,y \rangle$ pair at a time. To achieve this we model the learning task as follows: 

\begin{table}[!t]
	
	\centering
	\hspace{-0.6cm}
	\begin{tabular}{l|p{120pt}|p{123pt}|p{125pt}}
		\hline
		&The square is above the rectangle.&The square is below the rectangle. &The square is below the rectangle.\\
		$x$&The triangle is to the left of the square.&The triangle is to the right of the square.&The triangle is to the right of the square. \\
		&Is the rectangle to the right of the triangle?&Is the rectangle to the right of the triangle?&Is the triangle below the rectangle?\\\hline
		$y$&Yes& No & Yes\\\hline
		
	\end{tabular}
	\caption{A set of examples taken from the Task $17$ of bAbI   question answering dataset.}
	\label{table:dataset}
\end{table}

\begin{definition}[\bfseries{Inductive Logic Programming for Distinct Examples}]
	An ILP task for \textit{Distinct Examples} (denoted as $ILP^{DE}$) is a tuple $\langle B,M,D \rangle$, where $B$ is an Answer Set Program, called the background knowledge, $M$ defines the set of rules allowed in hypotheses (the hypothesis space) and $D$ is the dataset containing a series of context dependent examples $\langle E_1,E_2,..., E_n \rangle$. Here each $E_i$ is a tuple $\langle O_i,E_i^+, E_i^- \rangle$ where,  $O_i$ is a logic program, called \textit{observation} , $E^+$ is a set of positive ground literals and $E^-$ is a set of negative ground literals. A hypothesis $H$ is an inductive solution of $T$ (written as $H \in ILP^{DE}(B,M,D)$) \textit{iff},
	
	$$H\cup B \cup O_i \vdash E_i^+, ~\forall i=1...n $$
	$$H\cup B \cup O_i \nvdash E_i^-, ~\forall i=1...n $$  
\end{definition} 

In this formulation, each example $\langle O_i,E_i^+, E_i^- \rangle$ directly corresponds to an $\langle x,y \rangle$ pair and it takes into consideration that there are several distinct examples in a dataset, so there is no need to explicitly isolate them from each other. Table $2(b)$ shows the encoding of the running example in the format of $ILP^{DE}$. It turns out that the $ILP^{DE}$ task described here is a simplification of the \textit{Context-dependent Learning from Ordered Answer Sets task} proposed in \cite{law2016iterative}. However, to solve the \textit{Context-dependent Learning from Ordered Answer Sets task} the authors in \cite{law2016iterative} convert it to a standard ILP problem which creates the same scalability issue. 

\begin{table}[!htb]
	
	\centering
	
	\caption{ The $sample$ predicate is used to separate different examples. The constants $tri, rec, sq$ respectively denote triangle, rectangle and square. $holdsAt(rp(sq, rec, above),1)$ says that the square is above the rectangle at time point $1$.}	
	\begin{subtable}{.5\linewidth}
		\label{table:encoding_ilp}
		\begin{tabular}{|p{8pt}|l@{}|}
		\hline
		&$ans(X,no) \leftarrow not~ ans(X,yes), id(X).$\\
		\hline
		&$sample(1, holdsAt(rp(sq, rec, above),1)).$\\
		&$sample(1, holdsAt(rp(tri, sq, left),1)).$\\
		&$ans(1,yes)\leftarrow$ \\
		&$~~sample(1,holdsAt(rp(rec, tri, right),1)).$\\\cline{2-2}
		&$sample(2, holdsAt(rp(sq, rec, below),1)).$\\
		$B$&$sample(2, holdsAt(rp(tri, sq, right),1)).$\\
		&$ans2(yes)\leftarrow  $\\
		&$~~sample(2,holdsAt(rp(rec, tri, right),1)).$\\\cline{2-2}
		&$sample(3, holdsAt(rp(tri, sq, left),1)).$\\
		&$sample(3, holdsAt(rp(tri, sq, left),1)).$\\
		&$ans(3,yes)\leftarrow$\\&$~~ sample(3,holdsAt(rp(tri, rec, below),1)).$\\\hline
		$E^+$& \{$ans(1,yes)$,$ans(2,no)$,$ans(3,yes).$\}\\\hline
		$E^-$& \{$ans(1,no)$,$ans(2,yes)$,$ans(3,no).$\}\\\hline	
	\end{tabular}
	
	\caption{An ILP encoding of the problem in Table \ref{table:dataset} }
	\end{subtable}%
	\begin{subtable}{.5\linewidth}
	\begin{tabular}{|l|l|l|@{}}
		\hline
		&&$holdsAt(rp(sq, rec, above),1).$\\
		&$O_1$&$holdsAt(rp(tri, sq, left),1).$\\
		$E_1$&&$ans(yes)\leftarrow holdsAt(rp(rec, tri, right),1).$\\\cline{2-3}
		&$E_1^+$& \{$ans(yes)$\}\\\cline{2-3}
		&$E_1^-$& \{$ans(no)$\}\\\hline	
		&&$holdsAt(rp(sq, rec, below),1).$\\
		&$O_2$&$holdsAt(rp(tri, sq, right),1).$\\
		$E_2$&&$ans(yes)\leftarrow  holdsAt(rp(rec, tri, right),1). $\\\cline{2-3}
		&$E_2^+$& \{$ans(no)$\}\\\cline{2-3}
		&$E_2^-$& \{$ans(yes)$\}\\\hline	
		&&$holdsAt(rp(tri, sq, left),1).$\\
		&$O_3$&$holdsAt(rp(tri, sq, left),1).$\\
		$E_3$&&$ans(yes)\leftarrow holdsAt(rp(tri, rec, below),1).$\\\cline{2-3}
		&$E_3^+$& \{$ans(yes)$\}\\\cline{2-3}
		&$E_3^-$& \{$ans(no)$\}\\\hline		
	\end{tabular}
	\label{table:encding_cdl}
	\caption{An $ILP^{DE}$ encoding of the problem in Table \ref{table:dataset} }
	\end{subtable}%
	\label{table:encding}
	
\end{table}

It should be noted that any standard ILP problem $\langle B,M,{E^+,E^-} \rangle$ can be thought of as an $ILP^{DE}$ problem with only one example, $\langle \{\},M,\langle (B,E^+,E^-)\rangle \rangle$. Similarly any $ILP^{DE}$ task can be converted to an $ILP$ task. However, utilizing the `distinctness' property of the examples we can do better. The algorithm that we propose here roughly works as follows: Given an instance of the $ILP^{DE}$ task, it first finds a solution $H_1$ of $E_1$. Then it expands $H_1$ minimally to solve only $E_2$ and obtains $H_2$ . In the next iteration it again expands $H_2$ minimally to solve $E_1$ and it continues expanding until it finds a hypothesis that solves both $E_1$ and $E_2$. Next it starts with a solution of $\langle E_1,E_2\rangle$ and tries to expand it iteratively until it solves all of $E_1,E_2$ and $E_3$. The process continues until a hypothesis is found that explains all the examples. Section \ref{sec:i2xhail} describes the algorithm. We show that the algorithm is sound and complete when $H\cup B \cup O_i$ is \textit{stratified} for all $i=1,...,n$.
Our algorithm allows more control over the mode declarations (Section \ref{sec:background}) which can  lead to noticeable speed up in the search process.  We evaluate our algorithm on two popular datasets: 1) a question answering dataset published by Facebook AI Research \cite{weston2015towards} and 2) a handwritten digit recognition database \cite{lecun1998mnist}. To the best of our knowledge, no sound and complete ILP algorithm could learn from these two datasets. The work of \cite{mitra2016addressing} that learns from the bAbl dataset uses a modification of an existing ILP algorithm and the resulting algorithm is not complete. We discuss this further in section \ref{sec:exp}.

\section{Background}
\label{sec:background}
In this section, we describe the type of rules that our algorithm can deal with, the syntax of the mode declarations and the XHAIL algorithm which plays a crucial role in our algorithm.
 
\subsection{Answer Set Programming}
\label{subsec:asp}
An answer set program is a collection of rules of the form,
\begin{align*}
L_{0}  \leftarrow L_{1},...,L_m, \textbf{not } L_{m+1},..., \textbf{not } L_n
\end{align*}
where each of the $L_i$'s is a literal in the sense of a classical logic. Intuitively, the above rule means that if 
$L_{1},...,L_m$ are true and if $L_{m+1},..., L_n$ can be safely assumed to be false then 
$L_0$ must be true. The left-hand side of an ASP rule is called the \textit{head} and the right-hand side is called the \textit{body}. Predicates and ground terms in a rule start with a lower case letter, while variable terms start with a capital letter. We will follow this convention throughout the paper. A rule with no  \textit{head} is called a $constraint$. A rule with empty \textit{body} is referred to as a $fact$. The semantics of ASP is based on the stable model semantics of logic programming \cite{gelfond1988stable}. In this work, both the background knowledge $B$ and the solution $H$ are a collection of such ASP rules.

%\paragraph{Example}
%\begin{align}
%\nonumber holdsAt(relativeposition(X,Y, left), T) \leftarrow ~~~~~~
%\\\nonumber holdsAt(relativeposition(Z, X,above), T),
%\\holdsAt(relativeposition(Y, Z, right), T).~~ \label{rule_1}
%\end{align}
%The above rule represents the knowledge that $X$ is to the left of $Y$ if there is an object $Z$ such that $X$ is directly above $Z$ and $Z$ is to the right of $Y$. An answer set program $P$ containing the above rule and the facts and rules in $O_1$ from Table \ref{table:dataset} will entail $answer(yes)$ ($E^+_1)$.

\subsection{Mode Declarations}
\label{subsec:modes}
Given a set of positive examples $E^+$, negative examples $E^-$ and some background knowledge $B$, an ILP algorithm computes a set of rules $H$ so that $B \cup H \models E$. The rules in $H$  are often restricted with a language bias that is specified by a series of mode declarations $M$ \cite{muggleton1995inverse}. One can think of this as a way of injecting expert knowledge for the learning task. 

There are two types of mode declarations, namely \textit{modeh} declarations and \textit{modeb} declarations. A \textit{modeh(s)} declaration  (Table \ref{tab:mode8}) specifies a literal \textit{s} that can appear as the head of a rule in $H$. A \textit{modeb(s)} declaration (Table \ref{tab:mode8}) specifies a literal \textit{s} that can appear in the body of a rule. The argument \textit{s} is called \textit{schema} and comprises of two parts: 1) an \textit{identifier} for the literal and 2) a list of \textit{placemakers} for each argument of that literal. A \textit{placemaker} is either \textit{+type} (input), \textit{-type} (output) or \textit{\$type} (constant), where \textit{type} denotes the type of the argument. An answer set rule is in the hypothesis space defined by $M$ (call it $L(M)$) if and only if its head (resp. each of its body literals) is constructed from the schema \textit{s} in a \textit{modeh(s)} (resp. in a \textit{modeb(s)}) in $L(M)$) as follows:

\begin{itemize}
	\item[-] by replacing an output (-) placemaker by a new variable.
	\item[-] by replacing an input (+) placemaker by a variable that appears in the head or in a previous body literal and
	\item[-] by replacing a ground (\$) placemaker by a ground term.
\end{itemize}

Table \ref{tab:mode8} shows a set of mode declarations $M_{sample}$ that one can use to solve the example problem in Table \ref{table:dataset}.There is only one \textit{modeh(s)} declaration in $M_{sample}$, where the schema is \textit{holdsAt(relativeposition(+op1,+op1, \$direction), +time )}. Assuming that there are only four constants of type \textit{directions}, the set of possible head literals are:

\begin{equation*}
\left\{\begin{aligned}
&holdsAt(relativeposition(X,Y,left),T), \\
&holdsAt(relativeposition(X,Y,right),T), \\ 
&holdsAt( relativeposition(X,Y,above),T),\\ 
&holdsAt(relativeposition(X,Y,below),T)\\
\end{aligned}\right\}
\end{equation*}

Where X and Y are variables of type $op1$ and T has type \textit{time}. There are three \textit{modeb} declarations and they restrict additions of literals to the body as directed by their individual schema. Note that the following rule,
\begin{align*}
 holdsAt(relativeposition(X,Y, left), T) \leftarrow holdsAt(relativeposition(Z, X,above), T),\\
 holdsAt(relativeposition(Y, Z, right), T).
\end{align*}

\noindent is in $L(M_{sample})$, as the head is allowed by the $modeh$ (Table \ref{tab:mode8}) and the third \textit{modeb} (Table \ref{tab:mode8}) allows the addition of $holdsAt(relativeposition(Z, X,above), T)$ with $Z$ being an output (new) variable and the first \textit{modeb} allows the addition of $holdsAt(relativeposition(Y, Z, right), T)$, as all the associated variables $Y,~Z$ and $T$ have appeared before.

\begin{table}[htb]
	\centering
\begin{tabular}{l} 
	\hline\hline		
	\textit{\#modeh} \textit{holdsAt(relativeposition(+op1,+op1,\$direction),+time).}\\\hline
	\textit{\#modeb} \textit{holdsAt(relativeposition(+op1,+op1,\$direction),+time).}\\
	\textit{\#modeb} \textit{holdsAt(relativeposition(+op1,-op1,\$direction),+time).}\\
	\textit{\#modeb} \textit{holdsAt(relativeposition(-op1,+op1,\$direction),+time).}\\
	\hline\hline
\end{tabular}
\caption{Mode declarations for the problem of Table \ref{table:dataset}}
\label{tab:mode8}
\end{table}

Additionally, \textit{weights} can be assigned to $modeh$ and $modeb$ (written as \textit{\#modeh(s)=$W$}) and they express the cost that is involved when a mode declaration is used. The default weight for mode declarations is $1$. Existing implementations of the ILP algorithms, take only one set of mode declarations and thus all the $modeh$ declarations share the same set of \textit{modebs}. Our algorithm allows the user to provide \textit{modeh} specific \textit{modeb} declarations. This additional feature allows the user to provide more supervision in the search procedure and makes the search faster.

\subsection{XHAIL}
\label{s:xhail}
The XHAIL \cite{ray2009nonmonotonic} algorithm plays a crucial role in the algorithm that we present here. In this section, we describe various concepts and notations associated with the XHAIL algorithm.
Given an ILP task $ILP({B,M,E=\{E^+\cup E^-\}})$, XHAIL \cite{ray2009nonmonotonic} derives the hypothesis in three steps, namely the \textit{abductive} step, the \textit{deductive} step and the \textit{inductive} step. We will explain these steps with respect to the example $E_1$ from Table $2(b)$. The set $B$ contains the representation of $x_1$, denoted by $O_1$ and the set $E$ the contains annotations derived from $y_1$. $M$ is the set of mode declarations described in Table \ref{tab:mode8}. 

\subsubsection{Abductive Step}
In the first step XHAIL finds a set of ground (variable free) atoms $\triangle = \{\alpha_1,...,\alpha_n\}$ such that $B\cup \triangle \models E$, where each $\alpha_i$ is a ground instance of the \textit{modeh(s)} declaration atoms. For the running example there is only one \textit{modeh} declaration. Thus the set $\triangle$ can contain ground instances of only $holdsAt(relativeposition(X,Y,Z),T)$. In the following we show one possible $\triangle$ that meets the above requirement.

\begin{equation*}
\triangle = \left\{ 
\begin{aligned}
&holdsAt(relativeposition( rectangle,triangle,right),1) 
\end{aligned}
\right\}
\end{equation*}
\subsubsection{Deductive Step}
In the second step, XHAIL computes a clause $\alpha_i \leftarrow \delta_i^1...\delta_i^{m_i}$ for each $\alpha_i$ in $\triangle$, where $B\cup \triangle \models \delta_i^{j}, \forall 1\le i \le n, 1\le j\le m_i $ and each clause $\alpha_i \leftarrow \delta_i^1...\delta_i^{m_i}$ is a ground instance of a rule in ${L(M)}$. In the running example, $\triangle$ contains only one atom, $\alpha_1 = holdsAt(relativeposition(rectangle,triangle,\newline right),1) $ which is initialized to the head of the clause $k_1$. The body of $k_1$ is saturated by adding all possible ground instances of the literals in \textit{modeb(s)} declarations that satisfy the constraints mentioned above. There are two ground instances, $holdsAt(relativeposition(square,rectangle,above),1)$ and $holdsAt(relativeposition(triangle,square,left),1)$, of the literals in the \textit{modeb(s)} declarations and both of them can be added to the body as specified by $M$. In the following we show the set of ground clauses $K$ (called \textit{kernel}) constructed in this step  and their variabilized version ${K_\textit{v}}$ (called \textit{generalization}) that is obtained by replacing all input and output terms by variables.

\begin{equation*}
K = \left\{ 
\begin{aligned}
&holdsAt(relativeposition(rectangle,triangle, right),1) \\ 
& \hspace{0.5cm} \leftarrow holdsAt(relativeposition(square,rectangle,above),1),\\ 
& \hspace{1cm}holdsAt(relativeposition(triangle,square,left),1).
\end{aligned}
\right\}
\end{equation*}

\begin{equation*}
K_v = \left\{ 
\begin{aligned}
&holdsAt(relativeposition(X,Y,right),T) \\ 
& \hspace{20pt}\leftarrow holdsAt(relativeposition(Z,X,above),T),\\ 
& \hspace{35pt}holdsAt(relativeposition(Y,Z,left),T).\\
\end{aligned}
\right\}
\end{equation*}
\subsubsection{Inductive Step}
In this step XHAIL tries to find a compressive theory $H$ by selecting from $K_v$ as few literals as possible while ensuring that $B\cup H \models E$. For this example, working out this problem will lead to a unique solution,

\begin{equation*}
H = \left\{ 
\begin{aligned}
&holdsAt(relativeposition(X,Y,right),T). 
\end{aligned}
\right\}
\end{equation*}
which contains a single rule with empty body. In general, the compression process may lead to multiple options for $H$.

Let $\langle H_I,H_G,\triangle \rangle$ denote a solution returned by $XHAIL(B,M,E)$, where $H_G$ is the generalization computed from  $\triangle$ and $H_I$ is a compressed version of $H_G$ that solves $E$. It should be noted that there might be many choices for  $\triangle$ and correspondingly there might be many possible solutions $\langle H_I,H_G,\triangle \rangle $. In the following table, we define few notations which will be useful later. 

\begin{table}[htb]
	\label{tab:notation}
	\centering
	\begin{tabular}{l|p{9cm}} 
		\hline		
		\multicolumn{2}{c}{Notations}\\\hline
		XHAIL(B,M,E)& The set of all the solutions $\langle H_I,H_G,\triangle \rangle$ to the problem $P = ILP(B,M,E)$, where $H_I$ is minimal i.e. no compressed version of $H_I$ can solve $P$.\\
		$\triangle(B,M,E)$& \{$\triangle | \langle H_I,H_G,\triangle \rangle \in XHAIL(B,M,E)$ for some $H_I,H_G $\}.\\
		$H_G(B,M,E)$ & \{$H_G | \langle H_I,H_G,\triangle \rangle \in XHAIL(B,M,E)$ for some $\triangle,H_I $\}.\\
		$H_G(\triangle)$& The generalization computed from $\triangle$.\\\hline
		
	\end{tabular}

\end{table}

\section{Algorithm}
\textsc{XHAIL} can compute the solutions of $ILP(B_{E_1},M,\{E^+,E^-\}_{E_1})$. However how to compute the solutions of $ILP^{DE}(B,M,\langle E_1,E_2\rangle)$ without solving the standard Inductive Logic Programming task constructed from $E_1$ and $E_2$ (denoted by $ILP(B_{E_1,E_2},M,\{E^+,E^-\}_{E_1,E_2})$) ? This section addresses this question. Before that we define the following terms which will be needed for the discussion. 

\label{sec:i2xhail}

\begin{definition}{\bfseries{H\textsubscript{1} $\leq$ H\textsubscript{2}}}
	Two answer set programs $H_1 $ and $H_2$ are related by ``$\leq$'' (denoted as $H_1 \leq H_2$) if and only if $H_1 $ can be transformed into $H_2$ by either adding new rules to $H_1$ or by adding new literals in the body of the existing rules. 
	%One can verify that $\leq$ defines a partial order. 
\end{definition} 

\begin{definition}{\bfseries{Minimality}}
	A solution $H$ of $ILP(B,M, E)$ is \textbf{minimal} \textit{iff} $\not\exists H' < H$ in $L(M)$ that solves $ILP(B,M, E)$. 
\end{definition}

\begin{definition}{\bfseries{Distinctness}}
	A series of examples $E_i\langle O_i,E_i^+,E_i^-\rangle, i=1...n$  are said to be \textit{distinct} \textit{iff}, $\Delta(B\cup O_1\cup...\cup O_n ,M, \cup_{i=1}^n E_i^+, \cup_{i=1}^n E_i^-) = \{\cup_{i=1}^n \triangle_i | (\triangle_1,...,\triangle_n) \in  \Delta(B\cup O_1,M,E_1^+,E_1^-)\times ...\times \Delta(B\cup O_n,M,E_n^+,E_n^-)\} $. A series of examples $E_i\langle O_i,E_i^+,E_i^-\rangle, i=1...n$  are said to be \textit{mutually distinct} \textit{iff} all subsets of the examples are \textit{distinct}.
\end{definition}

Now consider the two examples $E_1$ and $E_2$ . Since $E_1$ and $E_2$ are \textit{distinct} examples constructed from two different $\langle x,y\rangle$ pairs,  by definition, $\Delta(B\cup O_1 \cup O_2,M,\cup_{i=1}^2 E_i^+, \cup_{i=1}^2 E_i^-) = \{\triangle_1 \cup \triangle_2 | (\triangle_1,\triangle_2) \in  \Delta(B\cup O_1,M,E^+_1,E_1^-) \times \Delta(B\cup O_2,M,E^+_2,E_2^-) \}$. Thus, for any solution $\langle H_I,H_G,\triangle \rangle$ of  $ILP(B\cup O_1 \cup O_2,M,\cup_{i=1}^2 E_i^+, \cup_{i=1}^2 E_i^-)$,  $\exists \triangle_1 \in \Delta(B\cup O_1,M,E^+_1\cup E_1^-)$ and $\exists \triangle_2 \in \Delta(B\cup O_2,M,E^+_2 \cup E_2^-)$ such that,

$$H_G(\triangle) = H_G(\triangle_1) \cup H_G(\triangle_2) \geq H_I $$

This property allows us to search for $H_I$'s without solving $ILP(B\cup O_1 \cup O_2,M,\cup_{i=1}^2 E_i^+, \cup_{i=1}^2 E_i^-)$ directly. The search procedure can be briefly described as follows: For any choice of $(\triangle_1,\triangle_2)$ pair, first find all the minimal $H \leq H_G(\triangle_1) \cup H_G(\triangle_2) $ that solves $E_1$  and then expand those minimally, with respect to $E_2$ and $E_1$ alternatively, until all the minimal $H_I$'s that solves both $E_1$ and $E_2$ are found. To find all the $H_I$ one simply needs to iterate over all possible $(\triangle_1,\triangle_2)$ pairs which can be computed from $ILP(B\cup O_1,M,E^+_1,E_1^-)$ and $ILP(B\cup O_2,M,E^+_2,E_2^-)$ individually. 

It should be noted that it is possible to have $H_G(\triangle') = H_G(\triangle'')$, even though $\triangle' \ne \triangle''$. Thus, the above search procedure can be optimized by iterating over pairs of generalizations instead of iterating over the abducibles. Another drawback of the above search procedure is that the search results of $(H^1_G(\triangle_1), H^2_G(\triangle_2))$ do not give any information for the search initiated on $(H^1_G(\triangle'_1), H^2_G(\triangle'_2))$. In every iteration it starts from scratch. However, if we remember the solutions of $ILP^{DE}(B,M,E_1)$, we can use those as lower bounds for finding the solutions of $ILP^{DE}(B,M,\langle E_1,E_2\rangle)$. This is because, if $H_I$ is a minimal solution of $ILP^{DE}(B,M,\langle E_1,E_2\rangle)$, then $H_I$ also solves $ILP^{DE}(B,M,E_1)$ and there exists a $\langle H^1_I,H^1_G,\triangle_1 \rangle ~\in ILP^{DE}(B,M,E_1)$ such that $H^1_I \leq H_I$. Thus, for the iteration $(H^1_G(\triangle_1), H^2_G(\triangle_2))$, one can search if some $H^1_I \leq H^1_G(\triangle_1)$ can be expanded by either expanding some rules in $H^1_I$ or by adding new rules from the remainder of  $H^1_G(\triangle_1)\cup H^2_G(\triangle_2)$  or both to solve $E_2$ along with $E_1$. Theorem \ref{th:1} formalizes this idea. 

\begin{theorem}
	\label{th:1}

For any solution $\langle H_I,H_G,\triangle \rangle$ of $ILP^{DE}(B,M,\langle E_1,..., E_n\rangle)$ there exists a solution $\langle H'_I,H'_G,\triangle' \rangle$ of $ILP^{DE}(B,M,\langle E_1,..., E_{n-1}\rangle)$ and a generalization $H''_G$ in $ILP^{DE}(B,M,E_n)$ such that, $H'_I \le H_I \le H'_G\cup H''_G $, when $H \cup B \cup O_i$ is  \textit{stratified} for any choice of $i \in \{1,...,n\}$ and $H \in \{H_G,H'_G,H''_G\}$. Here, $O_i$ is the \textit{observation} from $E_i$.
	$\blacksquare$
\end{theorem}

With this in mind, the algorithm for finding the solutions of $ILP^{DE}(B,M,\{E_1,E_2,..., E_n\})$ is described in Algorithm 1. The proof of the theorem is in Appendix A.

\begin{algorithm}
	
	\KwData{An instance of $ILP^{DE}({B,M,\{E_1,\ldots,E_n\}}$)}
	\KwResult{A solution to the problem}
	\tcc{initialize a stack with the solutions of $ILP(B,M,E_1)$}
	\textit{stack} = $XHAIL(ILP(B,M,E_1))$\; 
	\While{stack is not empty}{
		\tcc{pop the hypothesis from the top}
		$\langle H_I, H_G \rangle$ = $stack.pop()$\;
		
		\tcc{get an example $E_i$ such that $B\cup H_I \cup O_i \nvdash E_i^+$ or $B\cup H_I \cup O_i \vdash E_i^-$}
		$E_i$ = $nextUncoveredExample(H_I)$\;
		
		\tcc{No such example exists}
		\If{$E_i$ is null}
		{\tcc{found a solution} return $H_I$.} 
		\Else{ \tcc{Find expansions of $H_I$ that also solves $E_i$}
			$refinementsStack$ = $<>$ \;
			\tcc{support set denotes the set of examples from which $<H_i,H_G>$ is created}
			$supports$ = $supportSet(H_I)\cup \{E_i\}$\;
			
			\tcc{compute a set of lower bound-upper bound pairs for the search space.}
			$H_G(E_i)$ = $findGeneralizatons(B,M,E_i)$\;
			\ForEach{H in $H_G(E_i)$}{\textit{push} $\langle H_I,H_G\cup H \rangle $ to $refinementsStack$} 
			
			\While{refinementsStack is not empty}{
				\tcc{get a candidate lower bound-upper bound pair}
				$\langle H'_I,H'_G \rangle$ = $refinementsStack.pop()$\;
				
				\tcc{get an example from $supports$ that is not covered by $H'_I$}
				$E_j$ = $nextUncoveredExampleFromS(H'_I,$\ $supports)$\;
				
				\If{$E_j$ is null}{
					\tcc{if no such example exists then we found a solution to the subproblem. Push it to the $stack$.}
					\textit{push} $\langle H'_I,H'_G \rangle$ to $stack$; 
				}			
				
				\Else{
					\tcc{Expand $H'_I$ minimaly along $H'_G$ so that it covers $E_j$}
					$expansions$ = \textit{expandMinimal}($\langle H'_I, H'_G \rangle, B, E_j $)\;
					\tcc{Push all expansions in the $refinementsStack$ for further updates.}
					\ForEach{$\langle H''_I,H''_G \rangle$ in $expansions$}{\textit{refinementsStack}.\textit{push}($\langle H''_I,H''_G \rangle$)}
				}
			}
		}
	}
	
	\caption{$I^2XHAIL$}
	\label{algo}

\end{algorithm}

\subsection{Example}
In this subsection we describe how our algorithm computes a solution to the running example $ILP^{DE}(B,M,\langle E_1,E_2,E_3\rangle)$ from Table \ref{table:dataset}. Here $B$ contains all the constants of type $op1$, $direction$ and $time$ and $M$ is the one described in Table \ref{tab:mode8} .

\textbf{Initialization:} First the \textit{stack} is filled with the output from $XHAIL(B,M,E_1)$. In section 1, we have seen that the output contains only one tuple. The following block shows the content of the stack after initialization.The underlined part denotes $H_I$, where $H_G$ is the entire program.

%\FrameSep0pt
\begin{framed}
	\begin{equation*}
	\left.\begin{aligned}
	&\underline{holdsAt(relativeposition(X,Y,right),T)} \\ 
	&\hspace{20pt}\leftarrow holdsAt(relativeposition(Z,X,above),T),\\ 
	&\hspace{35pt}holdsAt(relativeposition(Y,Z,left),T).\\
	\end{aligned}\right.
	\end{equation*}
\end{framed}

\textbf{Iteration 1:} In iteration $1$, the hypothesis on the top (denoted as \textit{Top}$\langle H_I^{Top},H_G^{Top} \rangle$) of the stack is popped.  One can see that the hypothesis $H_I^{Top}$ does not cover $E_2$. So, the algorithm tries to find an expansion of it which solves $E_2$ and $E_1$ both. For that it first finds $H_G(B,M,E_2)$ and creates a new \textit{refinement stack} with lower bound ($H_I^{Top}$) - upper bound  ($H_G^{Top} \cup H_G^{Top}$) pairs as shown below:

\begin{framed}
	\begin{equation*}
	\left.\begin{aligned}
	&\underline{holdsAt(relativeposition(X,Y,right),T)} \\ 
	&\hspace{20pt}\leftarrow holdsAt(relativeposition(Z,X,above),T),\\ 
	&\hspace{35pt}holdsAt(relativeposition(Y,Z,left),T).\\
	\end{aligned}\right.
	\end{equation*}
\end{framed}

It may be noted that $H_G(B,M,E_2)$ is empty as $E_2$ does not contain any positive example, so the stack contains only and exactly the \textit{Top}. Next it pops the \textit{refinement stack}  and tries to find the minimal extensions of the \textit{Top} that covers $E_2$. There are two such minimal extensions , $H',H''$ and both of them are pushed to the  \textit{refinement stack}. 

\begin{equation*}
H' = \left\{
\begin{aligned}
&\underline{holdsAt(relativeposition(X,Y,right),T)} \\ 
&\hspace{20pt}\underline{\leftarrow holdsAt(relativeposition(Z,X,above),T)},\\ 
&\hspace{35pt}holdsAt(relativeposition(Y,Z,left),T).\\
\end{aligned}\right\}\\
\end{equation*}

\begin{equation*}
H'' = \left\{
\begin{aligned}
&\underline{holdsAt(relativeposition(X,Y,right),T)} \\ 
&\hspace{20pt}\leftarrow holdsAt(relativeposition(Z,X,above),T),\\ 
&\hspace{35pt}\underline{holdsAt(relativeposition(Y,Z,left),T)}.\\
\end{aligned}\right\}
\end{equation*}

The algorithm then goes on popping the top of the \textit{refinement stack}, say $H'$. Since $H'$ solves both $E_1$ and $E_2$  the condition on line $16$  of Algorithm 1 is satisfied and $H'$ is pushed into the main stack. Similarly, $H''$ is popped next and pushed to the main stack. At this point \textit{refinement stack} becomes empty and iteration $1$ exits as it has discovered all the minimal extensions of $Top$. The stack now contains $H''$ on top of $H'$. 

\textbf{Iteration 2:} In the next iteration the algorithm pops $\langle H''_I, H''_G \rangle$ which is currently at the top of the stack. The next problem that it does not solve is $E_3$. It then computes $H_G(B,M,E_3)$ which contain only one element,

\begin{equation*}
H''' = \left\{
\begin{aligned}
&holdsAt(relativeposition(X,Y,below),T) \\ 
&\hspace{20pt}\leftarrow holdsAt(relativeposition(Z,Y,below),T),\\ 
&\hspace{35pt}holdsAt(relativeposition(X,Z,right),T).\\
\end{aligned}\right\}\\
\end{equation*}

It then pushes $\langle H''_I, H''_G \cup H''' \rangle$ to the refinement stack and finds the minimal expansions of $ H''_I$ within the bound of  $H''_G \cup H'''$. There will be only one such expansion, $H^{final}$ which will then be pushed into the refinement stack and finally into the main stack. Since $H^{final}$ solves all three examples, the algorithms terminates returning $H^{final}$ as the solution.

\begin{equation*}
H^{final} = \left\{
\begin{aligned}
&holdsAt(relativeposition(X,Y,right),T) \\ 
&\hspace{20pt}\leftarrow holdsAt(relativeposition(Y,Z,left),T).\\
&holdsAt(relativeposition(X,Y,below),T) \leftarrow.\\
\end{aligned}\right\}
\end{equation*}

%\begin{equation*}
%ans_2 = \left\{
%\begin{aligned}
%&holdsAt(relativeposition(X,Y,below),T) \\ 
%&\hspace{20pt}\leftarrow holdsAt(relativeposition(Z,Y,below),T).\\
%&holdsAt(relativeposition(X,Y,below),T) \\ 
%&\hspace{20pt}\leftarrow holdsAt(relativeposition(Z,Y,below),T).\\
%\end{aligned}\right\}.
%\end{equation*}
%
%\begin{equation*}
%ans_3 = \left\{
%\begin{aligned}
%&holdsAt(relativeposition(X,Y,below),T) \\ 
%&\hspace{20pt}\leftarrow holdsAt(relativeposition(Z,Y,below),T).\\
%&holdsAt(relativeposition(X,Y,below),T) \\ 
%&\hspace{20pt}\leftarrow holdsAt(relativeposition(Z,Y,below),T).\\
%\end{aligned}\right\}.
%\end{equation*}
%
%\begin{equation*}
%ans_4 = \left\{
%\begin{aligned}
%&holdsAt(relativeposition(X,Y,below),T) \\ 
%&\hspace{20pt}\leftarrow holdsAt(relativeposition(Z,Y,below),T).\\
%&holdsAt(relativeposition(X,Y,below),T) \\ 
%&\hspace{20pt}\leftarrow holdsAt(relativeposition(Z,Y,below),T).\\
%\end{aligned}\right\}.
%\end{equation*}

\subsection{On the Minimality of the Solution}
The solution returned by algorithm \ref{algo} may not be minimal. This is because if $H_I$ is expanded minimally to $H'_I$ to solve a new example $E$, it does not ensure that $H'_I$ is minimal with respect to the relevant subproblem. An example of this is the following: $B = \{\}$, $E_1=\langle \{p.,b.,c.\},\{a\},\{\}\rangle$, $E_2=\langle \{b.\},\{\},\{a\}\rangle$, $E_3=\langle \{c.\},\{a\},\{\}\rangle$,  and $M = \{\# modeh~a, \# modeb ~b, \#modeb~ c, \#modeb~ p\}$. There are two solutions in $ILP^{DE}(B,M,\langle E_1,E_2 \rangle)$:  $H_1=\{a\leftarrow c.\}$ and $H2=\{a\leftarrow p.\}$. If $H_2$ is expanded first, it will produce $\{a\leftarrow p., a\leftarrow c. \}$ as the solution of $ILP^{DE}(B,M,\langle E_1,E_2, E_3\rangle)$ and since it covers all the examples, it will be returned as the solution.  However, only $\{a\leftarrow c.\}$ is sufficient to  cover $E_1,E_2,E_3$. Thus the output is not minimal. The minimal solution can be found by computing all the solutions to $ILP^{DE}(B,M,\langle E_1,E_2, E_3\rangle)$ and then discarding the ones which have a compressed version of it already in $ILP^{DE}(B,M,\langle E_1,E_2, E_3\rangle)$. However, algorithm \ref{algo} prefers efficiency over minimality and returns the first solution found.

\section{Related Work}
\label{sec:r_works}
In recent years the field of Inductive logic programming has seen major advancements in many of its areas. Different ILP algorithms have been proposed \cite{ray2009nonmonotonic,athakravi2013learning,law2014inductive,athakravi2015inductive,iled,kazmi2017improving,schuller2017best}. Researchers have analyzed various kinds of ``good'' rules that cannot be learned with the current definition of entailment (called ``cautious inference'')  and proposed an alternative to that, named as ``brave inference''. ILP Algorithms have thus been proposed that can do only ``brave inference'' \cite{otero2001induction} or both \cite{Sakama:2005,Sakama2009,law2015learning}. Efforts have also been made to learn answer set programs that not only contain Horn clauses but also choice rules and constraints \cite{law2015learning}. With these developments and the various systems that have been produced with these researches, people have successfully applied the paradigm of Inductive logic programming to various areas \cite{gulwani2015inductive,mitra2016addressing}. And with these exposures to different applications, several changes are being made to the paradigm of ILP.  

Recently \cite{law2016iterative} proposed context dependent learning for \textit{ordered} answer set programs. Due to lack of space we do not discuss learning ordered answer set programs here. Interested readers can refer to \cite{law2016iterative}. The definition of context dependent learning in this paper is an adaptation of their definition for standard ILP setting. It should be noted that even though the concept of  context depending learning was proposed in \cite{law2016iterative}, to solve the problem their method converts it to a standard ILP problem using choice rules. Here, we have made the first attempt to solve the problem in its original form. 

In this work, we deal with the situation where there are many small distinct examples $\{(x_1,y_1),..., \newline (x_n,y_n)\}$. Another situation where scalability is needed, is when there is a single but large example. Works in \cite{iled,DBLP:journals/corr/KatzourisAP17} talk about this situation. Our work is also related to the work in logical vision \cite{dai2015logical} that aims to learn symbolic representation of simple geometric concepts. 

%Also incremental algorithms for standard ILP problem which is useful when the the problem instance is big. However, they are not general and work for particular kind of ILP problem. Improvement by exploring "good quality" generalizations are also implemented. It is our future plan to incorporate "goodness" measure and approximate solutions. 

%Another approach that was reported in is to divide the dataset into small bunch and learn from the branches. It was an approach that worked for that dataset but finding those values are different and might not be possible always. In the experiment section we will show how $i^2xhail$ learns from the dataset on its own without any additional effort. 

\section{Experiments}
\label{sec:exp}
We have applied our algorithm on two datasets. They are discussed below:

\begin{table*}[!htb]
	\centering
	\begin{tabular*}{\textwidth}{ |p{115pt}|p{125pt}|p{125pt}|}
		\hline
		\textbf{Task 6: Lists/Sets} &\textbf{Task 17: Path finding}& \textbf{Task 10: Indefinite reasoning}\\\hline 
		Sandra picked up the football there.&The office is east of the hallway.&Fred is either in the school or the park.\\
		Sandra journeyed to the office.&The kitchen is north of the office.&Mary went back to the office.\\
		Sandra took the apple there.&The garden is west of the bedroom.&Bill is either in the kitchen or the park.\\
		Sandra discarded the apple.&The office is west of the garden.&Fred moved to the cinema.\\
		What is Sandra carrying?&How do you go from the kitchen to the garden?&Is Bill in the office?\\\hline
		
	\end{tabular*}
	\caption{Example question answering tasks from bAbI dataset }
	\label{tab:babi}
\end{table*}

\subsection{Question Answering} Recently a group of researchers from Facebook has proposed a question answering challenge \cite{weston2015towards} containing 20 different tasks. Table \ref{table:dataset} and \ref{tab:babi}  shows examples of such tasks. Each task contains $1000$ or more such stories in the training data. The goal is to build a system that uniformly solves all the tasks. 

The work of \cite{mitra2016addressing} has shown how Inductive logic programming can be used to solve the tasks. Their method can be summarized as follows: Given the input containing a story and a question, first translate it to an Answer Set Program using a natural language parser and some handwritten rules, then use some knowledge to answer the question. In the training phase, learn the necessary knowledge. They have used \textit{XHAIL} system to learn the knowledge. However, \textit{XHAIL} could not scale to the entire dataset. So they have divided the dataset. For each task their method takes a bunch of examples together, learns from the bunch using \textit{XHAIL}, adds the learned hypothesis back to the background knowledge and then takes the next bunch to learn from. Since knowledge learned from a group of examples is never updated again, they had to manually find a group size that will work for this dataset. The group size depended on the task and clearly it might happen that for some new task there does not exist a group size to which \textit{xhail} can scale. In this work, we reuse the dataset, their mode declarations and have found  that our algorithm can learn all the knowledge given the input $ILP^{DE}(B,M,D_{task})$, where $D_{task}$ contains all the $1000$ examples of a task. Table \ref{tab:result} shows the time it has taken, the number of rules learned for each task and the accuracy for each task. Our system has achieved the same accuracy as that of \cite{mitra2016addressing}.

\label{sec:exp}

\begin{table}[!t]
	\centering
	\hspace{-0.8cm}
	\begin{tabular}{ p{4cm} p{30pt} c p{0.5cm}}
		\hline
		TASK & Time & Rules & Acc\\\hline
		1: Single Supporting Fact&3 &10& 100\\
		2: Two Supporting Facts& 3 & 2 & 100\\
		3: Three Supporting facts& \_& \_&100\\
		4: Two Argument Relations & 2 &8& 100\\
		5: Three Argument Relations& 6& 20&100\\
		6: Yes/No Questions& \_& \_&100\\
		7: Counting& 5 & 14 &100\\
		8: Lists/Sets& 4& 8&100\\
		9: Simple Negation& 4& 13&100\\
		10: Indefinite Knowledge&9& 21& 100\\\hline
	\end{tabular}
	\quad
	\begin{tabular}{ l p{30pt} c p{0.5cm}}
		\hline
		TASK & Time & Rules & Acc\\\hline
		11: Basic Coreference& 4& 5&100\\
		12: Conjunction& \_& \_&100\\
		13: Compound Coreference& \_& \_&100\\
		14: Time Reasoning& 4& 4&100\\
		15: Basic Deduction& 4& 1&100\\
		16: Basic Induction& 4& 1&93.6\\
		17: Positional Reasoning& 4& 26&100\\
		18: Size Reasoning& 4& 4&100\\
		19: Path Finding& 17& 2&100\\
		20: Agent's Motivations& 2& 6&100\\\hline
	\end{tabular}
	\caption{Performance on the set of 20 tasks. The tasks for which training is not required is marked with `-'. \textit{Running time} is measured in \textit{minutes}. }
	\label{tab:result}
	
\end{table}	

\paragraph{Semantic  Parsing} We have done further experiments with the task of semantic parsing. We took all the unique sentences in the training dataset of \cite{weston2015towards} and the corresponding parse tree of the sentences and then trained an ILP system to do the conversion from scratch.
Table  \ref{tab:sptable} shows an example of this task. The training dataset contains 5458 such examples. Our system learned a collection of $165$ rules in $128$ minutes from the training data which accurately parsed all the sentences in the test data. 
%This shows how Inductive logic programming can be used to do semantic parsing. 
\begin{table}[!htb]
	\centering
	\begin{tabular}{ |l|}
		\hline
		\textbf{Sentence} \\\hline Daniel journeyed to the bathroom.\\\hline
		\textbf{ASP Representation $O_i$} \\\hline 
		index(1..5). lemma(1,daniel). pos(1,nn). lemma(2,journey).
		
		pos(2,vbd).
		lemma(3,to).\\
		
		pos(3,to).
		lemma(4,the).
		
		pos(4,dt).
		lemma(5,bathroom).		
		pos(5,nn).
		\\\hline
		\textbf{Positive Examples} $E_i^+$\\\hline 
		arg1(journey01,daniel), 
		arg2(journey01,bathroom) .\\\hline
		\textbf{Positive Examples} $E_i^-$\\\hline
		any possible output that is not in $E^+$.\\\hline
	\end{tabular}
	\caption{An example from the semantic parsing task. For each word in the sentence the representation contains its lemma and pos tag, which are obtained using Stanford parser .}
	\label{tab:sptable}

\end{table}

\subsection{Handwritten Digit Recognition} The MNIST dataset \cite{lecun1998mnist} contains images of handwritten digits. Each image is a $28\times 28$ matrix and is labeled with a number between $0$ to $9$ denoting the digit it represents. The value of a cell (pixel) in the matrix (image) ranges between 0 (black) to 255 (white) capturing the darkness at that point. In this experiment we use our ILP algorithm to learn rules that identifies digits. For that we represent the images in the following way:

\begin{enumerate}
	\item First, we divide all cell value by $255$ so that the value of each cell is in the range of $[0,1]$. 
	\item For each $4\times4$ non-overlapping submatrix we create a super-pixel whose value is the sum of the all the pixels in that region. This gives a $7\times7$ size matrix representation of the original image. Note that in this reduced matrix, each cell value ranges between $0$ to $16$.
	\item If the value of a super-pixel from the $7\times7$ matrix is less than $2$ we consider it to be in the \textit{off} state. If the value is more than or equal to $5$ we consider it be in the \textit{on} state. The original image is then described as two disjoint sets: 1) a set of positions where the state of the super-pixel is  \textit{off} and 2) another set where all the super-pixel are \textit{on}.
\end{enumerate}  

We learn rules on this representation. Each learned rule for a digit \textit{d} simply says, if the super-pixels in certain positions are \textit{off} and are \textit{on} for some other positions then the image represents the digit \textit{d}. The training data in the MNIST dataset contains a total of $60,000$ images with approximately $6,000$ images for each digit. To learn the rules for each digit we take all the examples of that digit and take equal amount of images that represent other digits and pass that to our algorithm. Table \ref{tab:result_mnist} shows the number of rules learned for each digit and the performance on the test data. Except for the digit $1$, it takes $160$ hours to learn the rules for each digit. 

\begin{table}[!htb]
	\centering

	\begin{tabular}{ l l c l}
		\hline
		Digit & \#Rules & \#Test Examples & Acc(\%)\\\hline
		0&3,021 &980& 60.91\\
		1& 444 & 1134 & 95.85\\
		2& 4,606&1032&32.95\\
		3 & 3,661 &1010& 49.80\\
		4& 3,416& 982&49.59\\\hline
	\end{tabular}
	\quad
	\begin{tabular}{ l l c l}
		\hline
		Digit & \#Rules & \#Test Examples & Acc(\%)\\\hline
		5&3,459 &891& 42.65\\
		6& 2,621 & 958 & 65.03\\
		7& 2,430& 1028&63.52\\
		8 & 3,237 &978& 54.50\\
		9& 2,382& 1009&69.18\\\hline
	\end{tabular}
	\caption{Performance on handwritten digit recognition tasks. For each digit, column 2 shows the numbers of rules learned, the number instances of that digit in the test set and the percentage of instances correctly classified.  }
	\label{tab:result_mnist}
	
\end{table}	

As the Table  \ref{tab:result_mnist} suggests the performance on handwritten digit recognition is quite poor in comparison to the state-of-the-art neural network classifier \cite{wan2013regularization} that achieves $99.79$\% accuracy on this dataset. The number of rules column in Table provides insights on this high error rates. Consider the example of digit $0$. If there are $5000$ instances of digit $0$ and the algorithm outputs $3,021$ rules that means the representation that we have chosen does not allow good generalization. However, the representation  seems to work quite well for the digit $1$.

An important lesson learned from this experiment is that even though it takes a small amount of time to perform a hypothesis refinement when finding a solution $H$ for  $\langle E_1,...,E_i \rangle$ from a solution of $\langle E_1,...,E_{i-1} \rangle$, the algorithm needs to verify if $H$ explains all of $\{E_1,...,E_i\}$ before it can proceed to the next iteration. If the size of $H$ is big (such as the case for digit recognition) and too many refinements are taking place then the algorithm spends a lot of time in the verification phase. An important future work will be to optimize this step  by identifying which examples could have been affected if a hypothesis goes through refinement. Nevertheless, the algorithm is able to output a solution and does not blow up when a problem of this size is given as input. The dataset associated with all the experiments and the learned rules are available at \url{https://goo.gl/k6AEEz}. All experiments were performed on an intel i7 machine with 12 GB RAM.

\label{sec:exp}

\section{Conclusion}
\label{sec:conclusion}
Earlier days of Artificial Intelligence have seen many handwritten rule based systems. Later those were replaced by better performing machine learning based systems. With the advancements of knowledge representation and reasoning languages, a natural question arises, ``if machines can learn logic programs, can it achieve better accuracy than existing statistical machine learning methods such neural networks?'' It should be noted that the system of \cite{mitra2016addressing} achieved better results than the existing deep learning models on the bAbI dataset. To further explore this possibility we need to focus on the task of learning of logic programs and need to develop systems that can learn from large datasets. In this paper, we have made an attempt towards that.  

\section*{Acknowledgments}
We are grateful to Stefano Bragaglia for making the code of XHAIL publicly available which is reused in the development of our system. We would also like to thank the reviewers for their insightful comments. This work has been supported by the NSF grant 1750082. 
	
\bibliographystyle{acmtrans}
\bibliography{i2xhailbib}

\begin{thebibliography}{}

\bibitem[\protect\citeauthoryear{Athakravi, Alrajeh, Broda, Russo, and
  Satoh}{Athakravi et~al.}{2015}]{athakravi2015inductive}
{\sc Athakravi, D.}, {\sc Alrajeh, D.}, {\sc Broda, K.}, {\sc Russo, A.}, {\sc
  and} {\sc Satoh, K.} 2015.
\newblock Inductive learning using constraint-driven bias.
\newblock In {\em Inductive Logic Programming}, pp.\  16--32. Springer, Cham.

\bibitem[\protect\citeauthoryear{Athakravi, Corapi, Broda, and Russo}{Athakravi
  et~al.}{2013}]{athakravi2013learning}
{\sc Athakravi, D.}, {\sc Corapi, D.}, {\sc Broda, K.}, {\sc and} {\sc Russo,
  A.} 2013.
\newblock Learning through hypothesis refinement using answer set programming.
\newblock In {\em International Conference on Inductive Logic Programming},
  pp.\  31--46. Springer.

\bibitem[\protect\citeauthoryear{Dai, Muggleton, and Zhou}{Dai
  et~al.}{2015}]{dai2015logical}
{\sc Dai, W.-Z.}, {\sc Muggleton, S.~H.}, {\sc and} {\sc Zhou, Z.-H.} 2015.
\newblock Logical vision: Meta-interpretive learning for simple geometrical
  concepts.
\newblock In {\em ILP (Late Breaking Papers)}, pp.\  1--16.

\bibitem[\protect\citeauthoryear{Gelfond and Lifschitz}{Gelfond and
  Lifschitz}{1988}]{gelfond1988stable}
{\sc Gelfond, M.} {\sc and} {\sc Lifschitz, V.} 1988.
\newblock The stable model semantics for logic programming.
\newblock In {\em ICLP/SLP}, Volume~88, pp.\  1070--1080.

\bibitem[\protect\citeauthoryear{Gulwani, Hernandez-Orallo, Kitzelmann,
  Muggleton, Schmid, and Zorn}{Gulwani et~al.}{2015}]{gulwani2015inductive}
{\sc Gulwani, S.}, {\sc Hernandez-Orallo, J.}, {\sc Kitzelmann, E.}, {\sc
  Muggleton, S.}, {\sc Schmid, U.}, {\sc and} {\sc Zorn, B.} 2015.
\newblock Inductive programming meets the real world.
\newblock {\em Communications of the ACM\/}~{\em 58,\/}~11, 90--99.

\bibitem[\protect\citeauthoryear{Katzouris, Artikis, and Paliouras}{Katzouris
  et~al.}{2015}]{iled}
{\sc Katzouris, N.}, {\sc Artikis, A.}, {\sc and} {\sc Paliouras, G.} 2015.
\newblock Incremental learning of event definitions with inductive logic
  programming.
\newblock {\em Machine Learning\/}~{\em 100,\/}~2-3, 555--585.

\bibitem[\protect\citeauthoryear{Katzouris, Artikis, and Paliouras}{Katzouris
  et~al.}{2017}]{DBLP:journals/corr/KatzourisAP17}
{\sc Katzouris, N.}, {\sc Artikis, A.}, {\sc and} {\sc Paliouras, G.} 2017.
\newblock Distributed online learning of event definitions.
\newblock {\em CoRR\/}~{\em abs/1705.02175}.

\bibitem[\protect\citeauthoryear{Kazmi, Sch{\"u}ller, and Sayg{\i}n}{Kazmi
  et~al.}{2017}]{kazmi2017improving}
{\sc Kazmi, M.}, {\sc Sch{\"u}ller, P.}, {\sc and} {\sc Sayg{\i}n, Y.} 2017.
\newblock Improving scalability of inductive logic programming via pruning and
  best-effort optimisation.
\newblock {\em Expert Systems with Applications\/}.

\bibitem[\protect\citeauthoryear{Law, Russo, and Broda}{Law
  et~al.}{2014}]{law2014inductive}
{\sc Law, M.}, {\sc Russo, A.}, {\sc and} {\sc Broda, K.} 2014.
\newblock Inductive learning of answer set programs.
\newblock In {\em European Workshop on Logics in Artificial Intelligence}, pp.\
   311--325. Springer, Cham.

\bibitem[\protect\citeauthoryear{Law, Russo, and Broda}{Law
  et~al.}{2015}]{law2015learning}
{\sc Law, M.}, {\sc Russo, A.}, {\sc and} {\sc Broda, K.} 2015.
\newblock Learning weak constraints in answer set programming.
\newblock {\em Theory and Practice of Logic Programming\/}~{\em 15,\/}~4-5,
  511--525.

\bibitem[\protect\citeauthoryear{Law, Russo, and Broda}{Law
  et~al.}{2016}]{law2016iterative}
{\sc Law, M.}, {\sc Russo, A.}, {\sc and} {\sc Broda, K.} 2016.
\newblock Iterative learning of answer set programs from context dependent
  examples.
\newblock {\em Theory and Practice of Logic Programming\/}~{\em 16,\/}~5-6,
  834--848.

\bibitem[\protect\citeauthoryear{LeCun}{LeCun}{1998}]{lecun1998mnist}
{\sc LeCun, Y.} 1998.
\newblock The mnist database of handwritten digits.
\newblock {\em http://yann. lecun. com/exdb/mnist/\/}.

\bibitem[\protect\citeauthoryear{Mitra and Baral}{Mitra and
  Baral}{2016}]{mitra2016addressing}
{\sc Mitra, A.} {\sc and} {\sc Baral, C.} 2016.
\newblock Addressing a question answering challenge by combining statistical
  methods with inductive rule learning and reasoning.
\newblock In {\em AAAI}, pp.\  2779--2785.

\bibitem[\protect\citeauthoryear{Muggleton}{Muggleton}{1991}]{muggleton1991inductive}
{\sc Muggleton, S.} 1991.
\newblock Inductive logic programming.
\newblock {\em New generation computing\/}~{\em 8,\/}~4, 295--318.

\bibitem[\protect\citeauthoryear{Muggleton}{Muggleton}{1995}]{muggleton1995inverse}
{\sc Muggleton, S.} 1995.
\newblock Inverse entailment and progol.
\newblock {\em New generation computing\/}~{\em 13,\/}~3-4, 245--286.

\bibitem[\protect\citeauthoryear{Otero}{Otero}{2001}]{otero2001induction}
{\sc Otero, R.} 2001.
\newblock Induction of stable models.
\newblock {\em Inductive Logic Programming\/}, 193--205.

\bibitem[\protect\citeauthoryear{Ray}{Ray}{2009}]{ray2009nonmonotonic}
{\sc Ray, O.} 2009.
\newblock Nonmonotonic abductive inductive learning.
\newblock {\em Journal of Applied Logic\/}~{\em 7,\/}~3, 329--340.

\bibitem[\protect\citeauthoryear{Sakama}{Sakama}{2005}]{Sakama:2005}
{\sc Sakama, C.} 2005.
\newblock Induction from answer sets in nonmonotonic logic programs.
\newblock {\em ACM Trans. Comput. Logic\/}~{\em 6,\/}~2 (April), 203--231.

\bibitem[\protect\citeauthoryear{Sakama and Inoue}{Sakama and
  Inoue}{2009}]{Sakama2009}
{\sc Sakama, C.} {\sc and} {\sc Inoue, K.} 2009.
\newblock Brave induction: a logical framework for learning from incomplete
  information.
\newblock {\em Machine Learning\/}~{\em 76,\/}~1 (Jul), 3--35.

\bibitem[\protect\citeauthoryear{Sch{\"u}ller and Kazmi}{Sch{\"u}ller and
  Kazmi}{2017}]{schuller2017best}
{\sc Sch{\"u}ller, P.} {\sc and} {\sc Kazmi, M.} 2017.
\newblock Best-effort inductive logic programming via fine-grained cost-based
  hypothesis generation.
\newblock {\em arXiv preprint arXiv:1707.02729\/}.

\bibitem[\protect\citeauthoryear{Wan, Zeiler, Zhang, Le~Cun, and Fergus}{Wan
  et~al.}{2013}]{wan2013regularization}
{\sc Wan, L.}, {\sc Zeiler, M.}, {\sc Zhang, S.}, {\sc Le~Cun, Y.}, {\sc and}
  {\sc Fergus, R.} 2013.
\newblock Regularization of neural networks using dropconnect.
\newblock In {\em International Conference on Machine Learning}, pp.\
  1058--1066.

\bibitem[\protect\citeauthoryear{Weston, Bordes, Chopra, and Mikolov}{Weston
  et~al.}{2015}]{weston2015towards}
{\sc Weston, J.}, {\sc Bordes, A.}, {\sc Chopra, S.}, {\sc and} {\sc Mikolov,
  T.} 2015.
\newblock Towards ai-complete question answering: a set of prerequisite toy
  tasks.
\newblock {\em arXiv preprint arXiv:1502.05698\/}.

\end{thebibliography}
\label{lastpage}

\newpage
\appendix
\section{Proof of Theorem \ref{th:1}}
\textbf{Theorem 1}

\noindent For any solution $\langle H_I,H_G,\triangle \rangle$ of $ILP^{DE}(B,M,\langle E_1,..., E_n\rangle)$ there exists a solution $\langle H'_I,H'_G,\triangle' \rangle$ of $ILP^{DE}(B,M,\langle E_1,..., E_{n-1}\rangle)$ and a generalization $H''_G$ in $ILP^{DE}(B,M,E_n)$ such that, $H'_I \le H_I \le H'_G\cup H''_G $, when $H \cup B \cup O_i$ is  \textit{stratified} for any choice of $i \in \{1,...,n\}$ and $H \in \{H_G,H'_G,H''_G\}$. Here, $O_i$ is the \textit{observation} from $E_i$.

\paragraph{Proof}

Recall that $\Delta(B,M,E)=$\{$\triangle | \langle H_I,H_G,\triangle \rangle \in XHAIL(B,M,E)$ for some $H_I,H_G $\}. We further define, 
\begin{center}
	\begin{tabular}{l}
		$\triangle(B,M,\langle E_1,...,E_n\rangle ) =\{(\triangle_1,\triangle_2,...,\triangle_n) | \triangle_i \in \triangle(B,M,E_i), \forall i=1..n\} $\\
		$H_G(\triangle = (\triangle_1,\triangle_2,...,\triangle_n)) = \cup_{i=1}^n H_G(\triangle_i)$
	\end{tabular}
\end{center}

\noindent Since $H_I$ is a solution to $ILP^{DE}(B,M,\langle E_1,E_2,..., E_{n-1}\rangle)$ and $H_I \cup B\cup O_i$ is assumed to be a stratified program, there is a unique set containing only ground instances of $modeh$ literals (abducible predicates), $\triangle^* = (\triangle_1^*,\triangle_2^*,...,\triangle_{n-1}^*)$ in $\triangle(B,M,\langle E_1,...,E_n\rangle)$ such that $\forall i \in {1,...,n-1}$,

\begin{enumerate}[label=\roman*]
	
	\item $B \cup O_i \cup H_I \vdash \triangle_i^*, $
	\item $\not\exists \triangle'_i. (\triangle'_i\in \triangle(B,M,E_i))\land ( B\cup O_i \cup H_I \vdash \triangle'_i) \land (\triangle_i^* \subset \triangle'_i).$
\end{enumerate}

Similarly, since $H_I$ is a solution to $ILP^{DE}(B,M,E_n)$ there is a unique $\bar\triangle$ such that,
\begin{enumerate}[label=\roman*]
	
	\item $B \cup O_n \cup H_I \vdash \bar\triangle, $
	\item $\not\exists \triangle'_n. (\triangle'_n\in \triangle(B,M,E_n))\land ( B\cup O_n \cup H_I \vdash \bar\triangle) \land (\bar\triangle \subset \triangle'_n).$
\end{enumerate}

$H_I$ is then bounded by $H_G(\triangle^*) \cup H_G(\bar\triangle)$. If this is not the case then $H_I$ has at least one rule whose body is not satisfied in any of the context provided by $B \cup O_i$, for all $i=1,...,n$. And hence $H_I$ cannot be minimal. Now consider the set $S$ containing all the minimal solution $\langle H_I',H_G',\triangle^* \rangle$ of $ILP^{DE}(B,M,\langle E_1,E_2,..., E_{n-1}\rangle)$ that can be obtained from $\triangle^*$. Let $H_I^*$ denote the set of all rules from $H_I$ that are satisfied in at least one of the context $B \cup O_i \cup H_I$, for $i=1...n-1$. Then, there must exist at least one $H_I' \in S$ such that $H_I' \leq H_I^*\leq H_I$. Otherwise, $H_I^*$ is a minimal solution of $ILP^{DE}(B,M,\langle E_1,E_2,..., E_{n-1}\rangle)$ that can be obtained from $\triangle^*$ but not in $S$. A contradiction. $\blacksquare$

\end{document}